\documentclass{article}

\usepackage{arxiv}

\usepackage[utf8]{inputenc} 
\usepackage[T1]{fontenc}    
\usepackage{hyperref}       
\usepackage{url}            
\usepackage{booktabs}       
\usepackage{amsfonts}       
\usepackage{nicefrac}       
\usepackage{microtype}      
\usepackage{lipsum}
\usepackage{graphicx}
\usepackage{subfigure}
\usepackage{amsmath}
\usepackage{array}
\usepackage{multirow}

\title{CF-CAM: Cluster Filter Class Activation Mapping for Reliable Gradient-Based Interpretability}

\author{
 Hongjie He \\
  School of Physics, Mathematics, and
Computing\\
  University of Western Australia\\
  Perth, Western Australia 6000, Australia \\
  \texttt{24462401@student.uwa.edu.au} \\
   \And
 Xu Pan \\
  School of Coumputer Science and Information Engineering\\
  Hefei University of Technology\\
  Hefei, Anhui 230009, China \\
  \texttt{2022217940@mail.hfut.edu.cn} \\
  \And
 Yudong Yao \\
  Department of Electrical and Computer Engineering\\
  Stevens Institute of Technology\\
  Hoboken, New Jersey 07030, USA \\
  \texttt{yyao@stevens.edu} \\
}

\begin{document}
\maketitle
\begin{abstract}
As deep learning continues to advance, the transparency of neural network decision-making remains a critical challenge, limiting trust and applicability in high-stakes domains. Class Activation Mapping (CAM) techniques have emerged as a key approach toward visualizing model decisions, yet existing methods face inherent trade-offs. Gradient-based CAM variants suffer from sensitivity to gradient perturbations due to gradient noise, leading to unstable and unreliable explanations. Conversely, gradient-free approaches mitigate gradient instability but incur significant computational overhead and inference latency. To address these limitations, we propose a Cluster Filter Class Activation Map (CF-CAM) technique, a novel framework that reintroduces gradient-based weighting while enhancing robustness against gradient noise. CF-CAM utilizes hierarchical importance weighting strategy to balance discriminative feature preservation and noise elimination. A density-aware channel clustering method via Density-Based Spatial Clustering of Applications with Noise (DBSCAN) groups semantically relevant feature channels and discard noise-prone activations. Additionally, cluster-conditioned gradient filtering leverages Gaussian filters to refine gradient signals, preserving edge-aware localization while suppressing noise impact. Experiment results demonstrate that CF-CAM achieves superior interpretability performance while enhancing computational efficiency, outperforming state-of-the-art CAM methods in faithfulness and robustness. By effectively mitigating gradient instability without excessive computational cost, CF-CAM provides a competitive solution for enhancing the interpretability of deep neural networks in critical applications such as autonomous driving and medical diagnosis.
\end{abstract}


\section{Introduction}
As deep learning technology continues to evolve, convolutional neural networks (CNN) have shown remarkable performance across various domains, including medical diagnosis \cite{jahmunah2022explainable}\cite{qi2025machine}\cite{altini2022ndg}\cite{zhang2023grad}\cite{talaat2024grad} and autonomous driving \cite{kolekar2022explainable}\cite{dworak2022adaptation}. Nevertheless, the "black box" characteristic of CNNs restricts the transparency of their decision-making processes. The Class Activation Mapping (CAM) technique offers a means to visualize and analyze the decision-making of CNNs, pinpointing key areas in the input image that influence these decisions. This aids users in gaining a better understanding of the model's decision-making process. The technique also assists in assessing the model's reliability, identifying potential biases, and boosting trust in intelligent systems. For instance, in medical analysis, CAM-based methods can precisely identify lesions in X-rays, aiding doctors in diagnosing; in autonomous driving, it can explain how the model perceives the external traffic environment, thereby refining the decision-making process and enhancing transparency.

Current advances in interpretability methods broadly fall into two categories: CAM-based and non-CAM approaches. CAM-based methods, such as Grad-CAM \cite{selvaraju2017grad} and its derivatives, leverage gradient propagation to generate saliency maps by weighting activation maps with class-specific gradients. Although Grad-CAM++ \cite{chattopadhay2018grad} mitigates multi-object localization challenges through second-order gradient terms and Smooth Grad-CAM++ \cite{omeiza2019smooth} enhances visual sharpness via noise-averaged gradient maps, these gradient-dependent methods remain susceptible to noise perturbations. To circumvent gradient instability, gradient-free CAM variants like Score-CAM \cite{wang2020score} and Ablation-CAM \cite{ramaswamy2020ablation} quantify feature importance through confidence-driven metrics or systematic activation ablation. However, these approaches often incur significant computational overhead. Parallel efforts, such as IS-CAM \cite{naidu2020cam} and Eigen-CAM \cite{muhammad2020eigen}, address efficiency limitations through stochastic integration or principal component decomposition, though trade-offs persist in balancing noise suppression, computational tractability, and nonlinear interpretability. Beyond CAM-based frameworks, model-agnostic methods like SHapley Additive exPlanations (SHAP) \cite{lundberg2017unified} and Local Interpretable Model-agnostic Explanations (LIME) \cite{ribeiro2016should} employed game-theoretic Shapley values or local surrogate modeling to explain diverse model architectures. While SHAP provides axiomatically consistent feature attribution, its sampling-based approximations struggle with high-dimensional data scalability. Similarly, LIME’s heuristic sampling and kernel weighting mechanism can introduce instability. These limitations underscore the need for interpretability techniques that integrate noise robustness, computational efficiency, and architectural flexibility across deep learning models.

Despite significant progress in interpretability technology research, there remain three fundamental challenges. Firstly, the faithfulness of explanations is crucial, as unstable visualization results may mislead users and affect their trust in the model's decisions. Secondly, inference time must be minimized, particularly in real-time scenarios where interpretable methods need to operate efficiently while maintaining model performance, quickly producing results and offering feedback. Finally, there are still shortcomings in the robustness of interpretations. For interpretable models, any minor disturbances should not alter the interpretation results, as inconsistency could reduce credibility in practical applications.

In order to tackle these challenges, we propose Cluster Filter Class Activation Map (CF-CAM) technique, a novel framework that synergizes channel clustering with gradient filtering to overcome these limitations. CF-CAM’s key innovations include:

\paragraph{Density-Aware Channel Clustering.} The Density-Based Spatial Clustering of Applications with Noise (DBSCAN) \cite{ester1996density} algorithm is applied to group feature channels into semantically coherent clusters while automatically discarding noise channels.

\paragraph{Cluster-Conditioned Gradient Filtering.} A Gaussian filtering mechanism is applied to gradients correspond with channels in each cluster, therefore suppress noise in the gradients while preserve semantic information.

\paragraph{Hierarchical Importance Weighting.} It differentially processes high-response dominant channels, clustered low-response channels, and excludes noise channels. This processing strategy eliminates spurious activations while preserving semantic consistency.

The remainder of this paper is organized as follows. Section II provides a review of related work. Our proposed method is described in detail in Section III. Section IV presents and analyzes the experimental results. Section V discusses future extensions and broader implications of CF-CAM. Finally, concluding remarks are provided in Section VI.

\section{Related Work}
\label{sec:headings}

\subsection{CAM Interpretability Methods}

CAM has become a core framework for the visualization of decision-making processes in deep neural networks. The methodology generates heatmap localization by associating feature maps with category scores, aiming to highlight discriminative regions. Early implementations like GAP-CAM \cite{zhou2016learning} faced architectural constraints due to their reliance on global average pooling layers, prompting subsequent research to focus on gradient dependency optimization and computational efficiency.

\subsubsection{Gradient-Based CAM Methods}

To address the structrual limitation of GAP-CAM, Grad-CAM \cite{selvaraju2017grad} introduced a gradient-based saliency mapping technique that computes the importance of feature maps by back-propagating category gradients to convolutional layers. However, this method suffers from gradient noise, which can make the results sensitive to small perturbations in input data and lead to unstable interpretability results. On the basis of Grad-CAM, Grad-CAM++ \cite{chattopadhay2018grad} introduced second derivative gradient terms to better capture the information of feature maps, especially in cases where multiple objects exist within the same image. However, this method still relies on gradients, making it susceptible to gradient interference or extreme values. Similarly, Smooth Grad-CAM++ \cite{omeiza2019smooth} enhances Grad-CAM++ by integrating the SmoothGrad \cite{smilkov2017smoothgrad} technique, which adds random noise to inputs and averages multiple gradient maps, resulting in visually sharper maps with improved object localization and capture. However, the impact of averaging gradients on preserving fine-grained details remains unexplored in this study. Layer-CAM \cite{jiang2021layercam} employs a hierarchical framework that aggregates feature activations across multiple convolutional layers to generate high-resolution saliency maps, achieving finer visual granularity and improved object localization precision. However, this approach imposes structural constraints as it requires compatible layer-wise feature representations, particularly in deeper networks where computational complexity escalates significantly. To address the theoretical limitations in existing gradient-based CAM methods, XGrad-CAM \cite{fu2020axiom} introduces sensitivity and conservation axioms through mathematical derivation to optimize feature map weighting. This axiom-driven approach enhances localization accuracy and interpretation reliability while maintaining gradient computation efficiency, though its performance remains dependent on gradient quality in deeper layers.

\subsubsection{Gradient-Free CAM Methods}

Ablation-CAM \cite{ramaswamy2020ablation} introduces a gradient-free approach that systematically ablates individual feature maps extracted from the final convolutional layer to quantify their importance through the resulting drop in class activation scores. While this method effectively avoids gradient saturation issues inherent in gradient-based methods like Grad-CAM, it requires $k+1$ forward passes (where $k$ is the number of feature maps), leading to increased computational costs. To address gradient-related issues such as noise and saturation in existing CAM methods, Score-CAM \cite{wang2020score} completely eliminates gradient dependence by introducing confidence-based activation weighting. Specifically, it determines each feature map's importance through its Class-specific Increase of Confidence (CIC) score, measured by forwarding masked inputs generated through activation upsampling. While this approach demonstrates excellent visual explainability, the requirement for individual forward passes per activation channel also introduces higher computational overhead compared to gradient-based alternatives. To address this problem, IS-CAM \cite{naidu2020cam} proposed a stochastic integration framework that iteratively averages activation impacts through multiple masked forward passes, effectively suppressing high-frequency noise in saliency maps while constraining computational costs to practical levels via optimized iteration control. Additionally, Eigen-CAM \cite{muhammad2020eigen} employs Singular Value Decomposition (SVD) on convolutional layer activations to compute their principal components, using the first eigenvector projection to generate CAMs. This approach eliminates gradient computation and feature weighting while maintaining compatibility with all CNN architectures. Although the method demonstrates robustness against classification errors and adversarial noise, its reliance on linear decomposition of activation features may limit the interpretation of nonlinear relationships learned by deep neural networks. Opti-CAM \cite{zhang2024opti} proposed an automated optimization strategy, enabling it to adapt dynamically across different network architectures, while using gradients indirectly for weight optimization rather than direct saliency map generation, thus reducing vulnerability to gradient instability.

\subsection{Non-CAM Interpretability Methods}

In addition to CAM-based approaches, several model-agnostic interpretability methods have been proposed to explain deep learning models in a broader context. SHAP \cite{lundberg2017unified} established a unified theoretical framework by leveraging cooperative game theory to provide feature importance scores that uniquely satisfy three desirable properties: local accuracy, missingness, and consistency. LIME \cite{ribeiro2016should} was designed to generate explanations by perturbing input samples and fitting local linear models, but its heuristic choices of kernel weighting and regularization parameters can lead to theoretically inconsistent explanations as shown in SHAP’s analysis \cite{lundberg2017unified}. This parameter sensitivity combined with the inherent randomness in sampling results in explanation instability. Anchor \cite{ribeiro2018anchors} extended LIME by identifying rule-based conditions that guarantee prediction stability. While improving robustness through formal coverage guarantees, its requirement for sufficient perturbation sampling increases computational complexity exponentially with feature dimensionality.

\section{Methodology}

\subsection{CF-CAM Architecture}

\begin{figure} 
    \centering
    \includegraphics[width=1\linewidth]{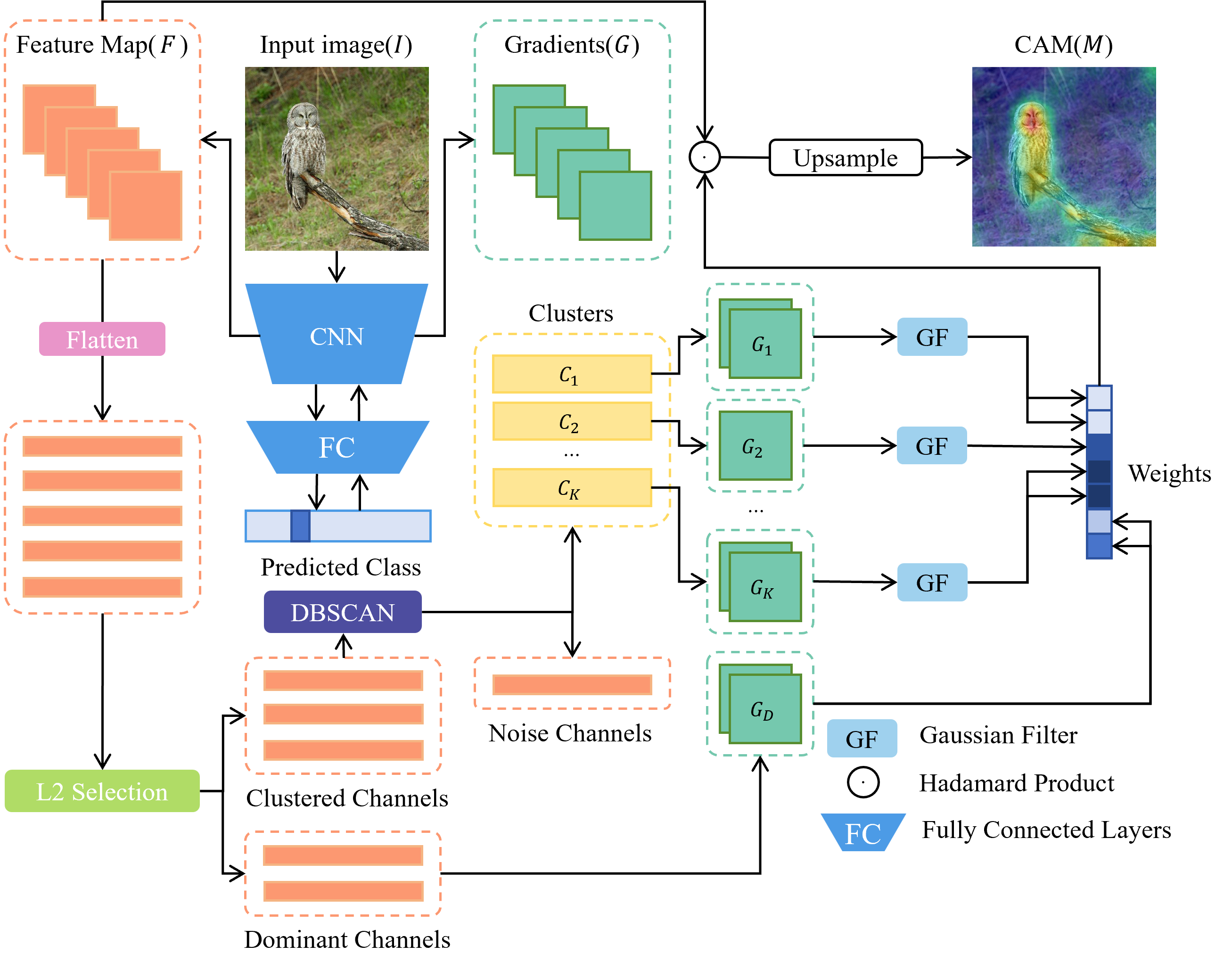}
    \caption{Overall pipeline of proposed CF-CAM.}
    \label{fig:overview}
\end{figure}

The proposed CF-CAM framework introduces a dual-stage approach to enhance visual interpretability in CNNs. As illustrated in Figure \ref{fig:overview}, our methodology comprises two synergistic stages: Channel Clustering Stage that groups semantically related feature channels, and Gradient Filtering Stage that refines gradient signals within each cluster.

Given an input image $I\in\mathbb{R}^{H\times W\times3}$, the process begins by extracting feature maps $F=\mathrm{CNN}(I)\in\mathbb{R}^{H^{\prime}\times W^{\prime}\times C}$  from a target convolutional layer of a pre-trained deep learning model, CF-CAM operates as follows: Initially, the Channel Clustering Stage employs a hierarchical clustering strategy to categorize the $C$ channels into three distinct groups: dominant channels $C_{dom}$ (high-response channels), clustered channels $C_{clu}$ (semantically similar low-response channels), and noise channels $C_{noi}$ (unclustered outliers) based on their L2 norm responses. The dominant channels are retained without further processing to preserve critical features, while the clustered channels are grouped using the DBSCAN algorithm whose parameters are tailored to the feature space. Subsequently, in the Gradient Filtering Stage, the gradients $\frac{\partial y^{class}}{\partial F_i}\in\mathbb{R}^{H^{\prime}\times W^{\prime}}$ corresponding to each channel are processed: for Dominant Channels, the raw gradients are used directly; for Clustered Channels, the mean gradient within each cluster is computed and refined using a differentiable bilateral filter $G_{e}$ to suppress noise while preserving edges. The refined gradients are then used to compute channel weights, which are normalized via a softmax function to ensure a balanced contribution across channels. Finally, the class activation map $M$, is generated by weighting the original feature maps with these normalized weights and applying a post-processing step $\mathcal{P}$ that includes ReLU activation and normalization. The total formulation of CF-CAM can be expressed as:

\begin{equation}
M=\mathcal{P}\left(\sum_{F_i\in\mathcal{V}}\omega_i\cdot F_i\right),
\end{equation}

where $\mathcal{V}=C_{dom}\cup C_{clu}$; $\omega_{i}$ represents the weight for channel $i$; $F_i\in\mathbb{R}^{H^{\prime}\times W^{\prime}}$ is the feature map of channel $i$. $P(x)=\frac{ReLU(x)}{max(ReLU(x))}$ is the post-processing operation, ensuring non-negative values and normalization to $[0,1]$. This design encapsulates the hierarchical processing and filtering mechanism, enabling CF-CAM to produce high-fidelity, noise-resistant heatmaps that highlight semantically meaningful regions in the input image.

\subsection{Channel Clustering Stage}

\begin{figure} 
    \centering
    \includegraphics[width=1\linewidth]{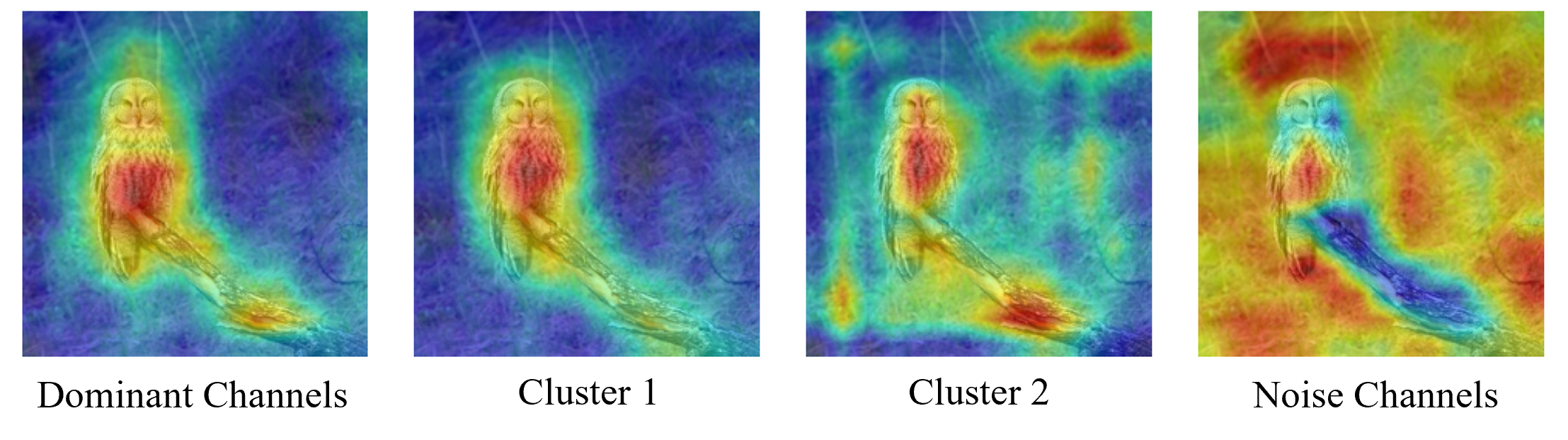}
    \caption{Visualization results of feature maps obtained from Channel Clustering Stage.}
    \label{fig:cluster}
\end{figure}

The Channel Clustering Stage in CF-CAM aims to group semantically related feature channels through density-based clustering to enhance gradient coherence. Given the extracted feature maps $F\in\mathbb{R}^{H\times W\times C}$ from the target convolutional layer, each channel’s spatial activation pattern is represented by flattening the spatial dimensions into a feature matrix $F'\in\mathbb{R}^{HW\times C}$. Initially, each channel’s L2 norm is computed and channels are separated using a threshold $\tau=Q_{p_1}(\{F_i\}_{i=1}^C)$, where the $Q_{p_1}(\cdot)$ function computes the $p_1$-th percentile of the input set. This produces dominant channels $C_{dom}=\{F_i|\|F_i\|_2\geq\tau,i\in[1,C]\}$ whose gradients are remained unprocessed and other remained channels which are going to be clustered. The L2 Selection process, on one hand, reduces computational overhead, as the clustering algorithms often produce significant computational overhead. On the other hand, the feature map values of high-response channels are typically concentrated in key regions, with a relatively sparse distribution, leading to significant spatial variations in gradient values. In contrast, the feature map values of low-response channels are more uniformly distributed, resulting in smaller gradient variations. The L2 selection process identifies high-response channels, ensuring that these channels, which are critical to model predictions, are processed separately, thus preventing unnecessary smoothing by the clustering algorithm.

To adaptively cluster other remained channels, the DBSCAN algorithm is employed with parameters dynamically derived from the data distribution. The pairwise Euclidean distance matrix $D\in\mathbb{R}^{C\times C}$ is computed between all channel vectors as:

\begin{equation}
D_{i,j}=\parallel F_i^{\prime}-F_j^{\prime}\parallel_2,
\end{equation}

where $F_i^{\prime}$ denotes the $i$-th row of $F^{\prime}$. The neighborhood radius $\epsilon$ is determined as the $p_2$-th percentile of the flattened distance matrix $D$. The minimum number of points $MinPts$ to form a dense region is set to $\max(2,\lceil0.01C\rceil)$. For each channel $F_i^{\prime}$, its $\epsilon-neighborhood$ is defined as $\mathcal{N}_\epsilon\left(F_i^{\prime}\right)=\{F_j^{\prime}|D_{i,j}\leq\epsilon\}$, and $F_i^{\prime}$ is classified as a core point if $|\mathcal{N}_\epsilon\left(F_i^{\prime}\right)|\geq MinPts$. Clusters expand iteratively by connecting density-reachable core points, yielding cluster labels $Lables=\{l_1,l_2,\cdots,l_C\}$, where $l_i (i\in[1,C])$ identifies noise channels excluded from subsequent processing. Valid clusters are defined as $\{\mathcal{C}_k|\mathcal{C}_k=\{i|l_i=k\},k\in\mathbb{N}^+\}$ , forming groups of channels with density-connected activation patterns. This stage ensures that only semantically coherent channels participate in local gradient refinement, while noise channels are discarded to mitigate spurious responses.

\subsection{Gradient Filtering Stage}

The Gradient Filtering Stage in CF-CAM operates on the clustered channels to refine gradients through semantic coherence constraints and noise suppression. Building on the cluster labels $Labels$ from the Channel Clustering Stage, this stage processes only valid clusters $\{\mathcal{C}_k\}$ to compute class-discriminative saliency maps. 

Our method preserves channel-specific information by individually filtering each channel's gradient. For a cluster $\mathcal{C}_k$ with $n_k$ channels, we collect the gradient values at each spatial position $(h, w)$ across all channels $i \in \mathcal{C}_k$, forming a sequence $\mathbf{G}_k(h, w) = [\frac{\partial y^{class}}{\partial F_i^{\prime}}(h, w)]_{i \in \mathcal{C}_k}$. Then we apply a 1D Gaussian filter to this sequence to produce smoothed gradient values, which can be represented as:

\begin{equation}
\widetilde{\mathbf{G}}_k(h, w) = \mathcal{F}_{\mathrm{Gaussian}}^{1\mathrm{D}}\left(\mathbf{G}_k(h, w)\right),
\end{equation}
where $\mathcal{F}_{\mathrm{Gaussian}}^{1\mathrm{D}}$ denotes the one-dimensional Gaussian filter. For each channel $i \in \mathcal{C}_k$, the channel-wise importance weight $\alpha_i$ is computed by spatially averaging its filtered gradient map:

\begin{equation}
\alpha_i = \frac{1}{H \cdot W} \sum_{h=1}^H \sum_{w=1}^W \widetilde{\mathbf{G}}_i(h, w).
\end{equation}

In the final step of the Gradient Filtering Stage, we apply softmax normalization across all valid channels to obtain normalized weights $\tilde{\alpha}_i$:
\begin{equation}
\tilde{\alpha}_i = \frac{\exp(\alpha_i)}{\sum_{j \in C_{clu}} \exp(\alpha_j)},
\end{equation}
where $\mathcal{V}$ represents the set of all valid channels (both dominant and clustered). These normalized weights ensure that the contributions of all channels are properly balanced in the final Class Activation Map.

\begin{equation}
\omega_i=
\begin{cases}
\frac{\mathrm{exp}(\widetilde{\alpha}_i)}{\sum_{F_j\in C_{clu}}\mathrm{exp}\left(\widetilde{\alpha}_j\right)+\sum_{F_j\in C_{dom}}\mathrm{exp}\left(\beta_j\right)} & \mathrm{for~}F_i\in C_{clu} \\
\frac{\mathrm{exp}(\beta_j)}{\sum_{F_j\in C_{clu}}\mathrm{exp}\left(\widetilde{\alpha}_j\right)+\sum_{F_j\in C_{dom}}\mathrm{exp}\left(\beta_j\right)} & \mathrm{for~}F_i\in C_{dom}
\end{cases}.
\end{equation}

The final CAM $M$ is generated by combining the normalized weights with the original feature activations:

\begin{equation}
M=\mathcal{P}\left(\sum_{F_i\in C_{clu}}\omega_i\cdot F_i+\sum_{F_i\in C_{dom}}\omega_i\cdot F_i\right)=\mathcal{P}\left(\sum_{i\in\mathcal{V}}\omega_i\cdot F_i\right).
\end{equation}

\subsection{Addresing Gradient Issues}

Wang et al. \cite{wang2020score} identified two critical flaws in gradient-based explanation methods: saturation, where activation functions like ReLU or Sigmoid cause vanishing gradients in semantically important regions, leading to underestimated channel weights; and false confidence, where noise or local gradient spikes in non-discriminative regions introduce unreliable high weights. To address the issue of saturation, CF-CAM leverages spatial-semantic clustering to group channels with similar activation patterns and computes refined gradients within each cluster. This compensates for individual channel saturation by aggregating preserved signals from non-saturated channels in the same cluster, ensuring semantically critical regions retain appropriate weights. For the false confidence issue, CF-CAM applies density-aware filtering—using DBSCAN to exclude sparse noise channels and enforcing cluster-level weight homogenization—which suppresses outlier gradients caused by transient perturbations. By assigning identical weights to all channels within a cluster, CF-CAM eliminates erratic weight spikes while preserving edge coherence through bilateral filtering. Together, these mechanisms help to resolve both saturation-induced underestimation and noise-driven overconfidence inherent in traditional methods.

\section{Experiments}

\subsection{Experiment Setup}

In this study, we evaluate the proposed CF-CAM in faithfulness, coherence, efficiency and noise robustness. The dataset we utilize in this study, Shenzhen Hospital X-ray Set \cite{jaeger2013automatic} , is a medical imaging dataset designed for tuberculosis detection, comprising 662 chest X-ray images with accompanying textual descriptions of pathological regions. The dataset is split into train (70\%), validation (20\%), and test (10\%) sets. The parameters of CF-CAM are set to: $p_1=75,p_2=10,\sigma_s=5.0,\sigma_r=0.1$. The parameter $m$ of the $\mathrm{AUC}_{del}$ and the $\mathrm{AUC}_{inc}$ are both set to 50. The parameter $K$ of average drop and average increase are both set to 50.

For the medical imaging experiments, we initially trained a fine-tuned ResNet-50 model adapted to the Shenzhen Hospital X-ray Set for binary classification (normal vs. tuberculosis), with the same target layer configuration(last convolutional layer of layer 4). We obtained an accuracy of 87.8\% for the ResNet-50 model. The baseline methods for comparison include Grad-CAM++, Score-CAM, Ablation-CAM, IS-CAM, and Opti-CAM, all implemented with consistent hyperparameters to ensure fairness.

All experiments are conducted on one NVIDIA RTX 3090 GPU with 24 GB of memory, using PyTorch 2.1.2 as the deep learning framework. Input images are resized to $224\times224$ and normalized using ImageNet \cite{deng2009imagenet} mean ([0.485, 0.456, 0.406]) and standard deviation ([0.229, 0.224, 0.225]).

\subsection{Evaluation Metrics}

We employ a comprehensive set of metrics to evaluate CF-CAM against baseline methods, focusing on localization accuracy, heatmap quality, and robustness.

Following previous works \cite{li2023bi,iqbal2023ad}, we utilize deletion and insertion curves to evaluate localization accuracy. The deletion curve measures prediction probability decay when progressively removing the most salient regions, while the insertion curve tracks probability increase when gradually adding salient regions to a blurred image. The area under both curves (AUC) provides a quantitative assessment, which can be represented as:

\begin{equation}
\mathrm{AUC}_{del}=\int_0^1f_{del}(m)dm,\mathrm{AUC}_{ins}=\int_0^1f_{ins}(m)dm,
\end{equation}

where $f_{del}(m)$ is the normalized prediction probability when $m$\% of the most salient pixels are masked, and $f_{inc}(m)$ is the normalized prediction probability when $m$\% of salient pixels are added.

To evaluate heatmap quality, we employ Average Drop (AD) and Average Increase (AI) metrics \cite{chattopadhay2018grad}. AD measures the relative decrease in prediction confidence when masking non-salient regions, while AI quantifies confidence improvement when preserving only salient regions. An effective CAM should exhibit low AD and high AI values.

For robustness evaluation, we calculate Structural Similarity Index Measurement (SSIM) \cite{wang2004image} and Mean Squared Error (MSE) between original heatmaps and their counterparts generated under gradient perturbation. Higher SSIM and lower MSE values indicate greater robustness to gradient noise.

\begin{table}[h]
\centering
\caption{Summary of Evaluation Metrics}
\begin{tabular}{|>{\centering\arraybackslash}p{1.5cm}|p{9cm}|>{\centering\arraybackslash}p{2.5cm}|}
\hline
\textbf{Metric} & \multicolumn{1}{c|}{\textbf{Description}} & \textbf{Interpretation} \\
\hline
$\mathrm{AUC}_{del}$ & Measures prediction decay as most salient regions are progressively removed & Lower is better \\
\hline
$\mathrm{AUC}_{inc}$ & Measures prediction increase as most salient regions are gradually added to a blurred image & Higher is better \\
\hline
$\mathrm{AD}$ & Quantifies prediction confidence decrease when non-salient regions are masked & Lower is better \\
\hline
$\mathrm{AI}$ & Measures prediction confidence improvement when only salient regions are preserved & Higher is better \\
\hline
SSIM & Structural similarity between original and perturbed heatmaps & Higher is better \\
\hline
MSE & Mean squared error between original and perturbed heatmaps & Lower is better \\
\hline
\end{tabular}
\end{table}

\subsection{Comparison Experiments}

To comprehensively evaluate the performance of our proposed CAM method, CF-CAM, we conduct a series of comparison experiments against state-of-the-art CAM techniques. These experiments aim to assess both the faithfulness and coherence of the generated heatmaps in highlighting regions relevant to the model’s predictions and the robustness of the methods that are gradient-based under gradient perturbations.

\subsubsection{Faithfulness, Coherence and Efficiency Evaluation}

\begin{table}[h]
    \centering
    \caption{Comparison of different CAM methods}
    \begin{tabular}{llccccc}
        \toprule
        \multicolumn{1}{c}{Category} & \multicolumn{1}{c}{Method} & \multicolumn{1}{c}{AD $\downarrow$} & \multicolumn{1}{c}{AI $\uparrow$} & \multicolumn{1}{c}{$\mathrm{AUC}_{del}$ $\downarrow$} & \multicolumn{1}{c}{$\mathrm{AUC}_{inc}$ $\uparrow$} & \multicolumn{1}{c}{$\mathrm{T}_{infer}$ $\downarrow$} \\
        \midrule
        \multirow{4}{*}{Gradient-free} & Ablation-CAM \cite{ramaswamy2020ablation} & 27.11\% & \textbf{7.79\%} & \textit{\textbf{0.5718}} & \textit{\textbf{0.7523}} & 8783.26 ms \\
        & Score-CAM \cite{wang2020score} & 27.85\% & 4.10\% & 0.6172 & 0.7227 & 9276.86 ms \\
        & IS-CAM \cite{naidu2020cam} & \textit{\textbf{27.09\%}} & 4.66\% & 0.6100 & 0.7242 & 174.73 ms \\
        & Opti-CAM \cite{zhang2024opti} & 29.42\% & 3.46\% & 0.6219 & 0.7177 & 135.61 ms \\
        \midrule
        \multirow{3}{*}{Gradient-based} & Grad-CAM++ \cite{chattopadhay2018grad} & 29.85\% & 2.60\% & 0.5871 & 0.7462 & \textbf{13.62 ms} \\
        & SG-CAM++ \cite{omeiza2019smooth} & 27.36\% & 3.94\% & 0.6131 & 0.7266 & \textit{\textbf{15.46 ms}} \\
        & CF-CAM (Ours) & \textbf{26.97\%} & \textit{\textbf{5.41\%}} & \textbf{0.5584} & \textbf{0.7586} & 712.33 ms \\
        \bottomrule
    \end{tabular}
    \label{tab:cam_comparison}
\end{table}

We evaluate the faithfulness of CF-CAM and some baseline CAM methods on the Shenzhen Hospital X-ray Set. We adopt four widely recognized metrics: AD, AI, $\mathrm{AUC}_{del}$, $\mathrm{AUC}_{inc}$ and inference time $\mathrm{T}_{infer}$. The $\mathrm{T}_{infer}$ is tested under an image resolution of $224\times224$ and a batch size of $1$.

The results of this evaluation are summarized in Table \ref{tab:cam_comparison}. The best results are shown in bold, while the second-best results are in bold and italic style. From the table, we observe significant variations in performance across different CAM methods. Our proposed CF-CAM achieves the lowest AD value of 26.47\%, outperforming existing approaches such as Grad-CAM++ (29.85\%), Score-CAM (27.85\%), and the recent Opti-CAM (29.42\%). This indicates that masking non-salient regions identified by CF-CAM results in the smallest degradation in model confidence, validating its precision in highlighting critical features. Moreover, CF-CAM also achieves second highest AI value of 5.41\%, surpassing methods such as Ablation-CAM (7.79\%) and IS-CAM (4.66\%). The results demonstrate the outstanding performance of CF-CAM in faithfulness.

The superiority of CF-CAM is further demonstrated in deletion and insertion AUC metrics. It achieves the lowest $\mathrm{AUC}_{del}$ (0.5584), significantly better than Grad-CAM++ (0.5871), Score-CAM (0.6172), and IS-CAM (0.6100), which reflects its ability to remove irrelevant regions without disrupting model predictions. Simultaneously, CF-CAM attains the highest $\mathrm{AUC}_{inc}$ (0.7586), outperforming all baselines, including Ablation-CAM (0.7523) and Smooth Grad-CAM++ (0.7266). This highlights its robustness in retaining discriminative regions critical for accurate predictions.

Notably, while Ablation-CAM and IS-CAM show competitive AD and AI values, CF-CAM’s consistent dominance across all four metrics underscores its holistic advantages. The results suggest that CF-CAM not only identifies salient regions more effectively but also aligns better with the model’s decision logic, as evidenced by its balanced performance in confidence preservation and spatial coherence.

From the last column of the table, it can be observed that CF-CAM achieves a balanced trade-off between computational efficiency and interpretation quality, with its inference time of 712.33 ms being 8564.53 ms and 8070.93 faster than traditional gradient-free methods like Score-CAM and Ablation-CAM. Our framework maintains real-time practicality while delivering superior localization accuracy as demonstrated in previous experiments.

\begin{figure} 
    \centering
    \includegraphics[width=1\linewidth]{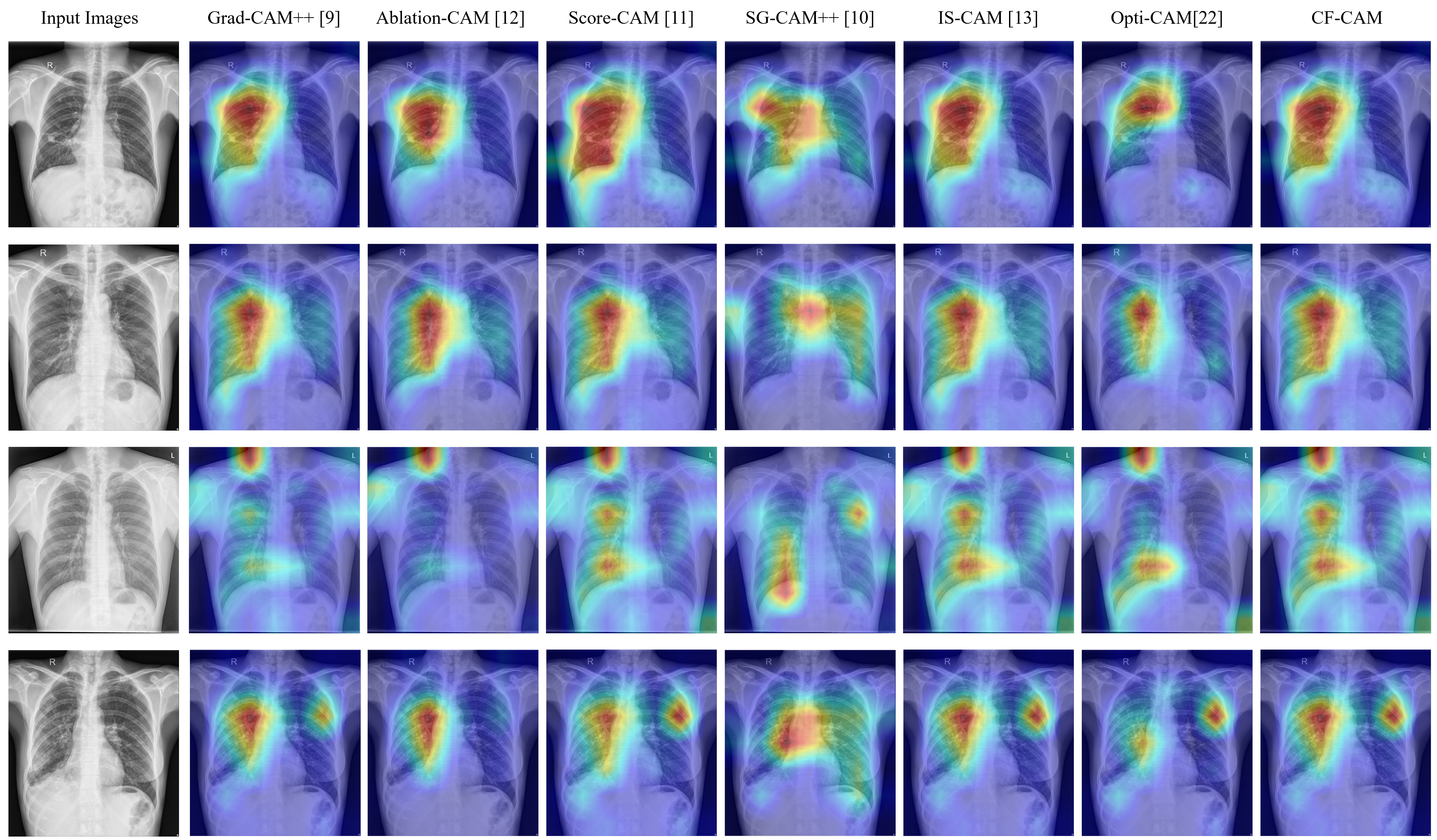}
    \caption{CAMs generated by different methods for four clinical cases from the Shenzhen Hospital X-ray dataset.}
    \label{fig:com1}
\end{figure}

To evaluate the clinical relevance of CAM-generated saliency maps, we visualize heatmaps for four representative chest X-ray cases from the dataset, each accompanied by radiologist-provided "Clinical Reading" annotations. As shown in Figure \ref{fig:com1}, which present heatmaps from Grad-CAM++, Ablation-CAM, Score-CAM, and our proposed CF-CAM. The clinical diagnoses for the cases (left to right) are: (1) right pulmonary tuberculosis (PTB) with fibrous changes, (2) left secondary PTB with right pleural effusion, (3) right secondary PTB, and (4) bilateral PTB.

For Case 1, CF-CAM localizes high-activation regions to fibrotic lesions in the right lung, aligning with radiological markers of reticular opacities and traction bronchiectasis. In contrast, Smooth Grad-CAM++ exhibits diffuse activations extending into healthy parenchyma. Case 2 demonstrates CF-CAM’s ability to resolve multifocal pathology: distinct activation peaks highlight the lesion area in both side of the lungs. Smooth Grad-CAM++ generates aberrant high-activation regions in the right lung field, likely attributable to gradient saturation artifacts in fibrotic tissue, whereas Opti-CAM fails to produce discernible activations in the left upper lobe—a critical discrepancy given the documented left secondary PTB diagnosis characterized by cavitary lesions. All baseline CAM methods exhibit spurious high-activation regions near the cervical spine in Case 3, a phenomenon attributable to the ResNet-50 backbone’s attention bias toward non-pathological high-contrast edges. CF-CAM successfully suppresses these confounding signals while localizing true pathological regions in the right lower lobe, whereas Ablation-CAM fails to activate corresponding areas. In Case 4, both Smooth Grad-CAM++ and Ablation-CAM demonstrate insufficient activation across bilateral lung fields, failing to localize diffuse pathological regions corroborated by clinical annotations. CF-CAM successfully labeled lesion areas bilaterally in the lungs.

\subsubsection{Gradient Perturbation Robustness Evaluation}

\begin{figure} 
    \centering
    \includegraphics[width=1\linewidth]{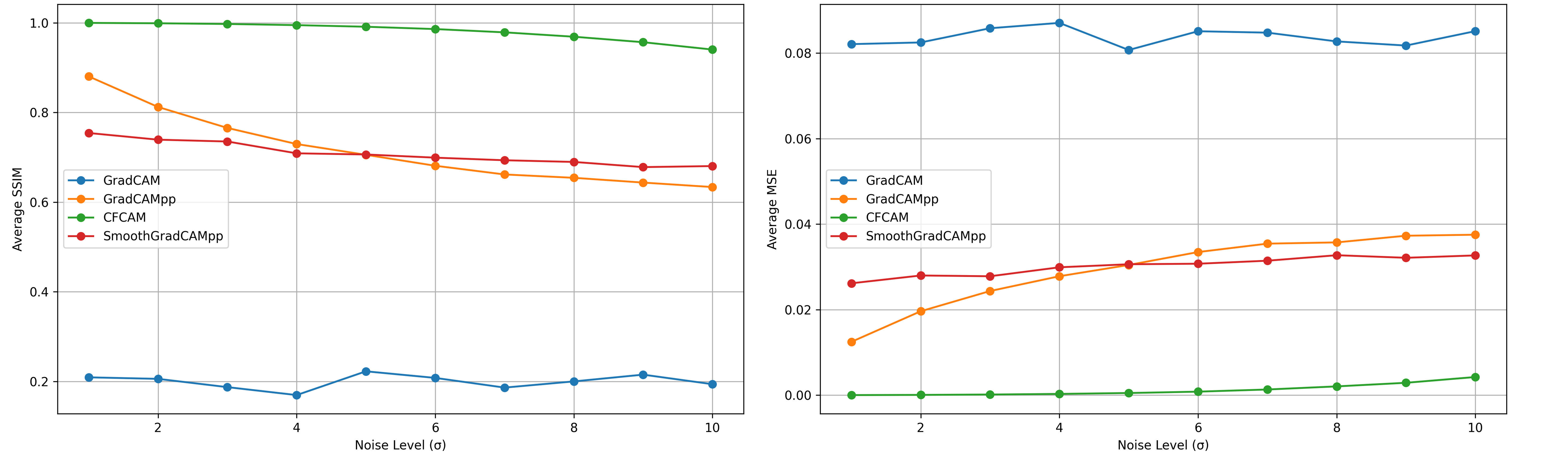}
    \caption{Average SSIM and MSE vs. Noise Level for Different CAM Methods based on gradient.}
    \label{fig:mse_ssim}
\end{figure}

In real-world application scenarios, the input images for deep learning models often contain noise. This not only challenges the generalization capabilities of the models but also impacts the performance of interpretation algorithms, particularly those based on gradients. Here, noise in the features can result in noisy gradients, thereby skewing the interpretation outcomes. 
To evaluate the robustness of gradient-based CAM methods against noise perturbations, we conduct a comprehensive experiment comparing GradCAM, GradCAM++, Smooth GradCAM++, and our proposed CF-CAM. These methods are selected because they all rely on gradient information to generate heatmaps, making them particularly susceptible to gradient perturbations. In this experiment, we simulate noise by adding Gaussian noise with varying standard deviations to the gradients obtained from the last convolutional layer of a ResNet-50 model.

The robustness of each CAM method is quantified using two metrics: SSIM and MSE. For each image in the dataset, we generate two heatmaps: one using the original gradients and another using the perturbed gradients. SSIM measures the perceptual similarity between the original and noisy heatmaps, with values closer to 1 indicating higher robustness, while MSE quantifies the pixel-wise numerical deviation, where lower values reflect greater stability. We compute the average SSIM and MSE across all test images for each noise level, providing a comprehensive assessment of robustness.

The robustness curves in Figure \ref{fig:mse_ssim} illustrate the average SSIM and MSE of Grad-CAM, Grad-CAM++, Smooth Grad-CAM++ (SG-CAM++), and CF-CAM across noise levels $\sigma$ from 1 to 10, with Figure 5’s left subplot showing SSIM declining as noise increases and the subplot on the right showing MSE rising correspondingly. CF-CAM, represented by the green line, consistently maintains the highest SSIM and the lowest MSE, demonstrating superior robustness to the gradient noise perturbation. Grad-CAM++ and SG-CAM++ exhibit similar trends, with SSIM values starting around 0.8 and dropping to approximately 0.65, and MSE values increasing from about 0.02 to 0.035, showing moderate robustness. Grad-CAM, depicted by the blue line, performs the worst, with SSIM fluctuates around 0.2 and MSE always above 0.08. In conclusion, CF-CAM consistently exhibits the best resistance to gradient noise, GradCAM++ and SG-CAM++ demonstrate comparable noise resistance that is lower than CF-CAM but higher than GradCAM, while GradCAM shows the lowest noise robustness among the methods evaluated.

\begin{figure} 
    \centering
    \includegraphics[width=1\linewidth]{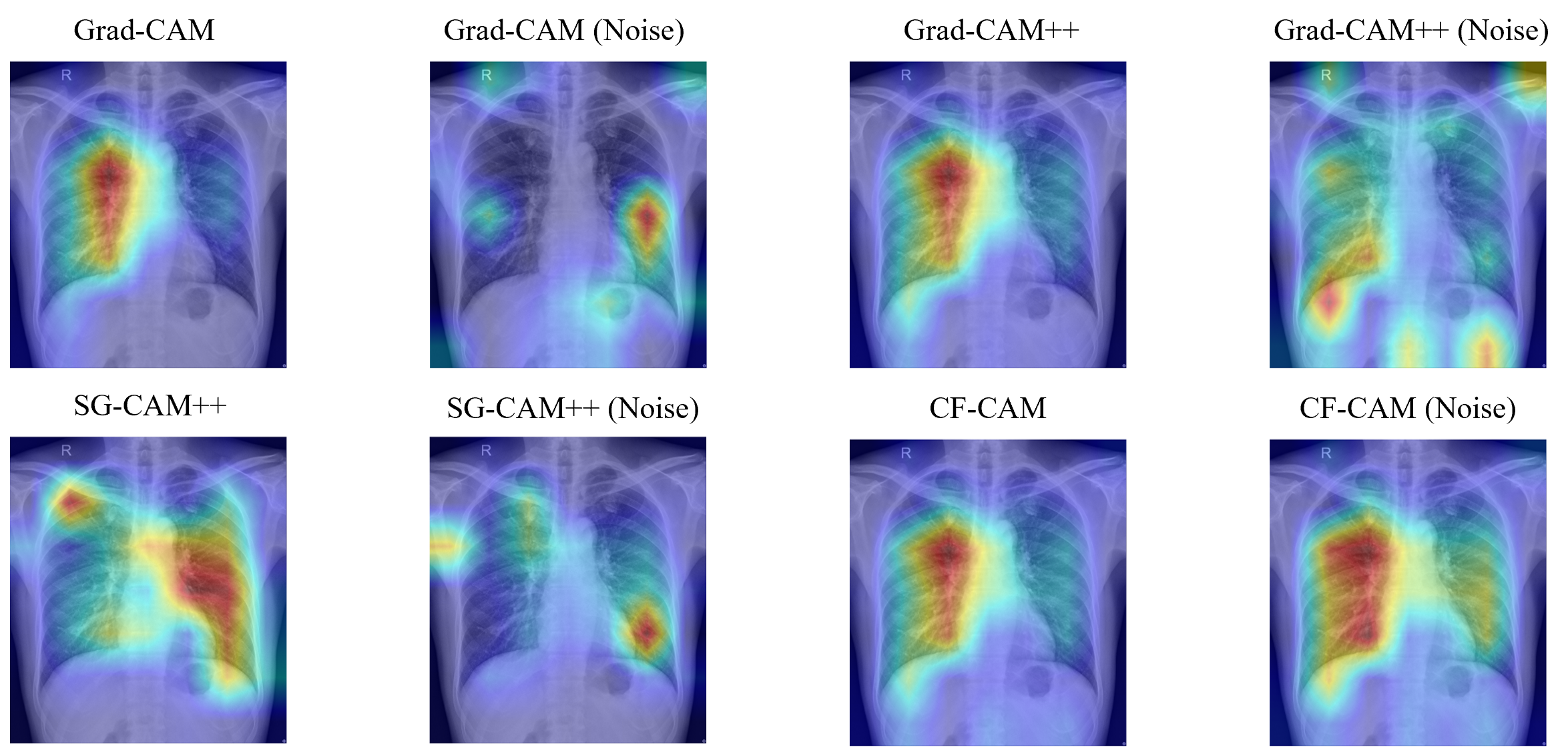}
    \caption{Original and Noisy heatmaps of Gradient-Based CAM Methods on a image randomly chosen from the Shenzhen Hospital X-ray Set. ($\sigma=5$)}
    \label{fig:noise}
\end{figure}

Figure \ref{fig:noise} demonstrates the output of four gradient-based methods before and after noise is added to the gradients. The original heatmaps of Grad-CAM, Grad-CAM++, and SG-CAM++ highlight specific regions in the lungs, but their noisy counterparts exhibit significant shifts and dispersion, indicating sensitivity to noise. In contrast, CF-CAM’s original heatmap focuses on a well-defined region in the right lung, and its noisy version remains largely unchanged, preserving the highlighted area with minimal deviation, underscoring CF-CAM’s superior robustness to gradient noise compared to the other methods, which show notable degradation.

\subsection{Ablation Study}

\subsubsection{Experiments on L2 Selection}
To assess the impact of the L2 Selection mechanism in our CF-CAM framework, we conducted an ablation study by comparing the performance of CF-CAM with and without L2 Selection. In this experiment, We implemented a variant of CF-CAM, which skips the L2 Selection step and directly applies DBSCAN clustering to all channels.

\begin{table}[h]
\centering
\caption{Ablation study results comparing CF-CAM with and without L2 Selection.}
\begin{tabular}{lccc}
\toprule
Method & AD $\downarrow$ & AI $\uparrow$ & \textbf{$\mathrm{T}_{infer}$ $\downarrow$} \\
\midrule
w/o L2 Selection & \textit{\textbf{33.19\%}} & \textit{\textbf{3.14\%}} & \textit{\textbf{1126ms}} \\
CF-CAM & \textbf{26.47\%} & \textbf{5.41\%} & \textbf{712ms} \\
\bottomrule
\end{tabular}

\label{tab:ablation_l2_with_timing}
\end{table}

The results in Table~\ref{tab:ablation_l2_with_timing} highlight the effectiveness of L2 Selection in the CF-CAM framework. With L2 Selection enabled, the CF-CAM achieves a lower AD value of 26.47\%, a higher AI value of 5.41\% and a lower $\mathrm{T}_{infer}$ of 712ms. It can be concluded that the L2 Selection mechanism efficiently preserved the key regions in the feature map and reduces inference time.

\subsubsection{Experiments on Clustering algorithm}

To evaluate the impact of the clustering algorithm choice in our CF-CAM framework, we conducted an ablation study comparing the performance of CF-CAM using different clustering methods: K-Means, DBSCAN, Spectral Clustering, and Gaussian Mixture Models (GMM). In this experiment, we replaced the default DBSCAN clustering in CF-CAM with each of these alternatives while keeping all other components unchanged.

\begin{table}[h]
\centering
\caption{Ablation study results comparing different clustering algorithms in the CF-CAM framework.}
\begin{tabular}{lcc}
\toprule
Clustering Method & AD $\downarrow$ & AI $\uparrow$ \\
\midrule
K-Means & 27.07\% & \textit{\textbf{5.17\%}} \\
Spectral Clustering & 27.13\% & 5.03\% \\
GMM & \textit{\textbf{27.06\%}} & 5.03\% \\
DBSCAN & \textbf{26.47\%} & \textbf{5.41\%} \\
\bottomrule
\end{tabular}

\label{tab:ablation_clustering}
\end{table}

The results presented in Table~\ref{tab:ablation_clustering}, demonstrate the superiority of DBSCAN within the CF-CAM framework. DBSCAN achieves the lowest AD value of 26.47\% and the highest AI value of 5.41\%, outperforming the other clustering methods. The results also show the effectiveness of the noisy channel elimination mechanism of DBSCAN.

\section{Discussion}

In this study, we introduce CF-CAM, a novel interpretability method designed to address the critical challenges of faithfulness, computational efficiency, and robustness in interpreting neural networks’ decision-making processes. CF-CAM leverages density-aware channel clustering, cluster-conditioned gradient filtering, and hierarchical importance weighting to overcome limitations of existing methods, such as gradient-dependent noise amplification and computational inefficiency of gradient-free approaches. The density-based clustering mechanism automatically separates semantically coherent channels from noise, while Gaussian filtering of gradient ensembles within clusters ensures edge-aware localization and suppresses spurious activations. The hierarchical processing strategy—retaining dominant channels, refining clustered channels, and discarding outliers—strikes a balance between preserving discriminative features and enhancing interpretability consistency.

Experimental results on the Shenzhen Hospital X-ray Set demonstrate CF-CAM’s superior performance compared to state-of-the-art CAM variants, particularly in medical imaging scenarios requiring precise lesion identification. Quantitative metrics confirm its high faithfulness to model decisions, with improved coherence in highlighting relevant image regions. Runtime analysis further validates its computational tractability, requiring only a single forward-backward pass, unlike gradient-free methods that demand multiple iterations. Robustness tests reveal that CF-CAM maintains consistent explanations under input perturbations, addressing a key limitation of gradient-sensitive methods. These results underscore CF-CAM’s potential to enhance trust in intelligent systems, such as medical diagnostics, where precise and reliable interpretability is critical.

Despite its strengths, CF-CAM has limitations that warrant further exploration. The method’s reliance on density-based clustering may be sensitive to hyperparameter settings, particularly in networks with highly variable channel distributions. Future work will focus on extending CF-CAM to multi-modal architectures and evaluating its applicability in real-time safety-critical systems, such as autonomous driving. Further theoretical analysis of the interplay between channel clustering granularity and semantic interpretability could refine its adaptability across diverse network depths and tasks.  By bridging the gap between theoretical robustness and practical efficiency, CF-CAM lays a foundation for developing trustworthy artificial intelligence systems in domains where interpretability is paramount.

\section{Conclusion}

This paper presents CF-CAM, a novel method for enhancing the interpretability of neural networks. By addressing challenges in faithfulness, computational efficiency, and robustness, CF-CAM provides reliable and efficient visualization of decision-making processes. Experimental results validate its effectiveness, particularly in medical imaging. CF-CAM establishes a foundation for trustworthy Artificial Intelligence systems, with potential applications in critical domains requiring transparent decision-making.

\bibliographystyle{unsrt}  
\bibliography{references}

\end{document}